\pdfoutput=1
\documentclass[11pt]{article}

\usepackage[utf8]{inputenc} % From ACL template
\usepackage{times} % From ACL template
\usepackage{latexsym} % From ACL template
\usepackage[dvipsnames]{xcolor} % For color names
\usepackage{amsmath} % Extended math typesetting
\usepackage{amssymb} % Extended math symbols
\usepackage{mathtools} % More mathtools
\usepackage{annotate-equations} % Help annotate equations
\usepackage{pifont} % More symbols (also needed for checkmark and x-mark symbol macros)
\usepackage{truncate} % Help truncate text
\usepackage{enumitem} % allows setting leftmargin on \begin{enumerate} or \begin{itemize}
\usepackage{csquotes} % Inline and display quotes
\usepackage{xurl} % For \url
\usepackage[dont-mess-around]{fnpct} % Automatically ensures footnotes go after the punctuation mark
\usepackage{soul} % For text highlighting
\usepackage{ragged2e} % allows \centering
\usepackage{float} % allows the [H] on \begin{figure}[H]

\usepackage{caption}
\usepackage{subcaption} % For subfigures / captions: https://tex.stackexchange.com/a/122329

\usepackage{graphicx} % allows \includegraphics

\usepackage{array}
\usepackage{longtable}
\usepackage{booktabs}
\usepackage{multirow}
\usepackage{arydshln}
\usepackage{tabularx}
\usepackage{tabulary}
\usepackage{makecell}

\usepackage{tikz,pgfplots}
\pgfplotsset{compat=1.17}

\usepackage{xinttools} % Useful to populate table data with a loop
\usepackage{ifthen} % If-then in LaTeX

\usepackage{lipsum}

\usepackage{todonotes}

\usepackage[T1]{fontenc}
\newcommand{\cmmnt}[1]{}

\newcommand{\mainfig}{
    \begin{figure}[ht!]
        \centering
        \includegraphics[width=150px]{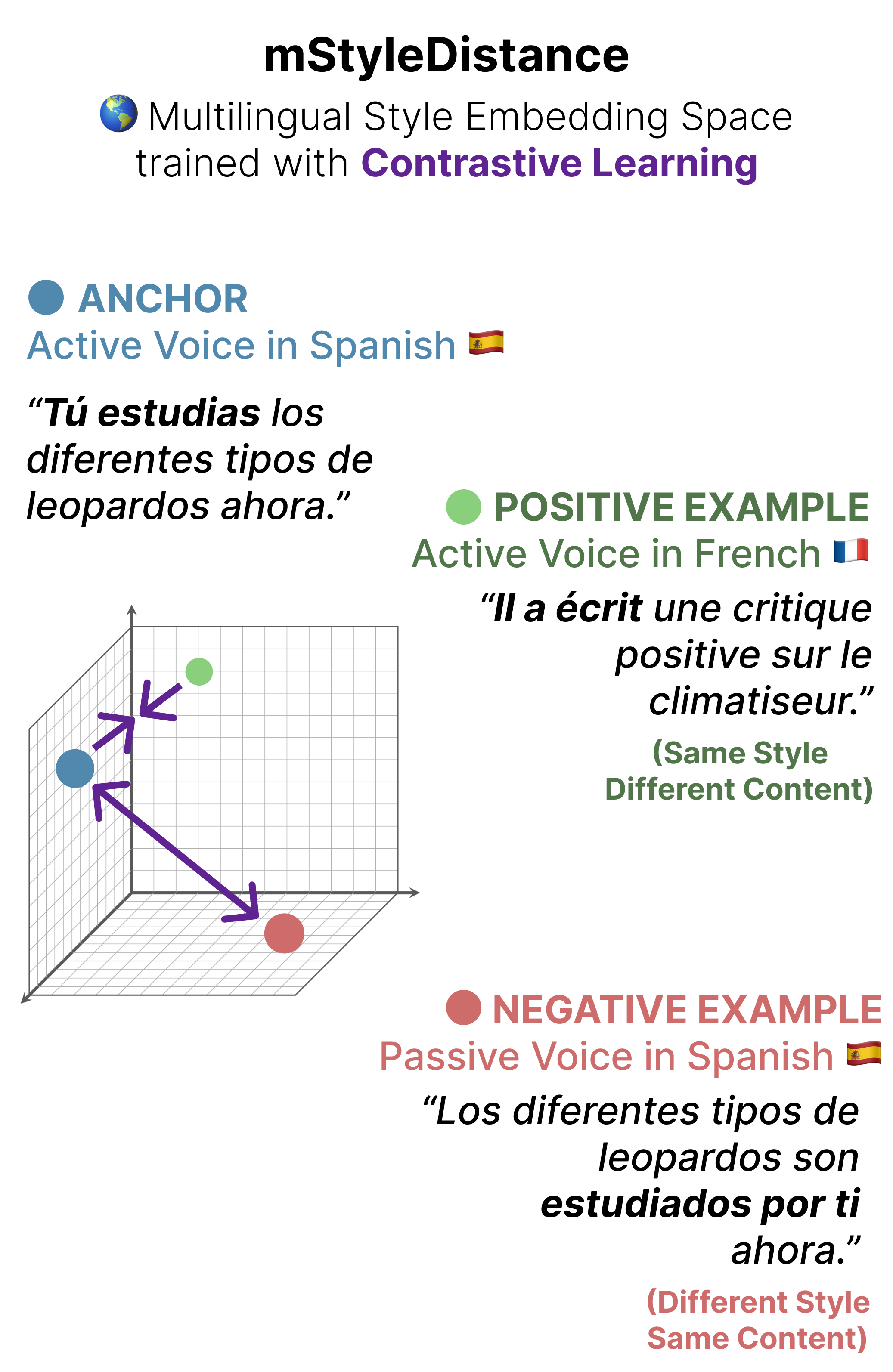}
        \caption{\textsc{mStyleDistance} is trained using contrastive learning from synthetic positive and negative examples of \textasciitilde 40 style features in 9 languages to form both multilingual and cross-lingual training triplets.}
        \label{fig:main}
    \end{figure}
}

\newcommand{\mturkinterfacefig}{
    \begin{figure}[H]
        \centering
        \begin{subfigure}{0.85\linewidth}
            \centering
            \includegraphics[width=\linewidth]{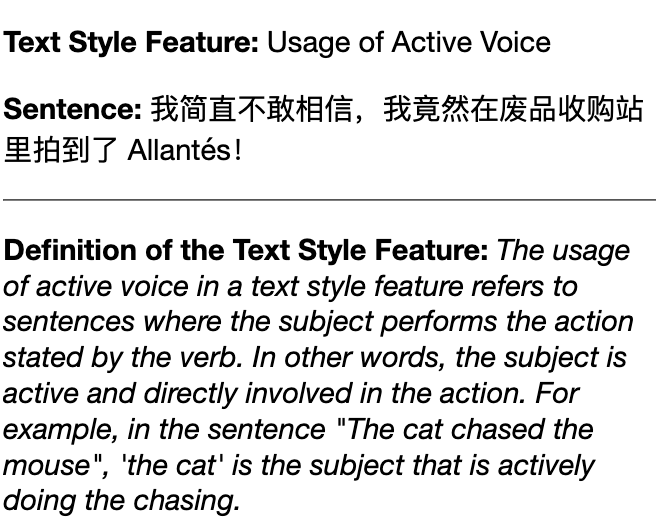}
            \label{fig:mturkinterface1}
        \end{subfigure}
        \vspace{0.1cm}
        \begin{subfigure}{\linewidth}
            \centering
            \includegraphics[width=0.85\linewidth]{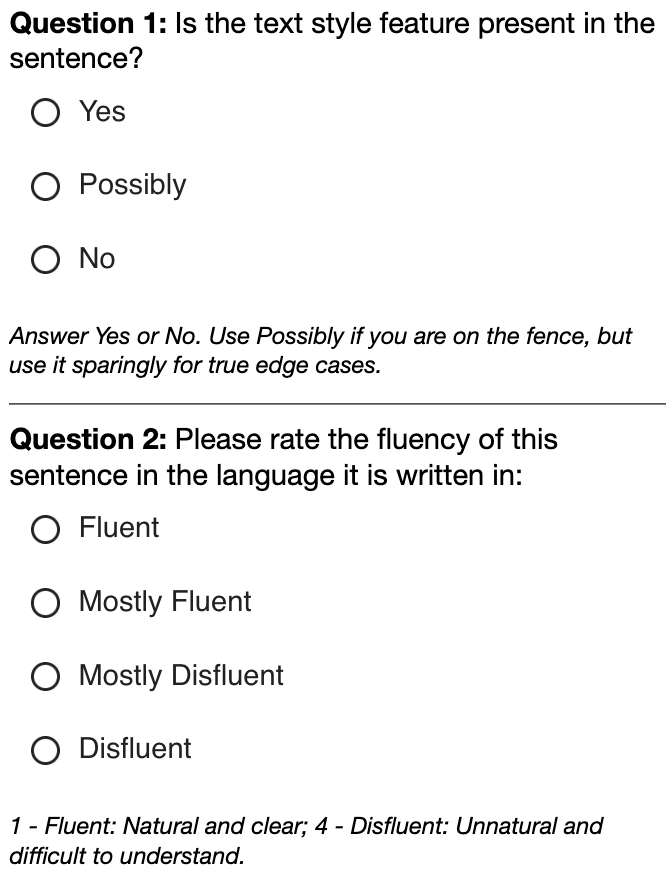}
            \label{fig:mturkinterface2}
        \end{subfigure}
        \caption{Instances from the annotation interface.}
        \label{fig:mturkinterface}
    \end{figure}
}

\newcommand{\stelevaltable}{
    \begin{table*}[!h]
    \centering
    \large
    \scalebox{0.75}{
    \begin{tabular}{l*{5}{c}}
    \toprule[\heavyrulewidth]
    \textbf{Model} 
    & \textbf{Simplicity} 
    & \textbf{Formality} 
    & \textbf{Toxicity} 
    & \textbf{Positivity} 
    & \textbf{Formality (cross-lingual)} \\ \midrule
    \textbf{\citet{styleemb}} 
    & 0.23 
    & 0.63 
    & 0.19 
    & 0.23 
    & 0.45 \\ 
    \textbf{\textsc{StyleDistance}} 
    & 0.21 
    & 0.67 
    & 0.15 
    & 0.18 
    & 0.49 \\ 
    \textbf{xlm-roberta-base} 
    & 0.12 
    & 0.16 
    & 0.09 
    & 0.07 
    & 0.19 \\ 
    \textbf{LISA} 
    & 0.15 
    & 0.09 
    & 0.09 
    & 0.21 
    & 0.27 \\ 
    \textbf{\textsc{mStyleDistance}} 
    & \textbf{0.36} 
    & \textbf{0.71} 
    & \textbf{0.37} 
    & \textbf{0.30} 
    & \textbf{0.53} \\ 
    \bottomrule[\heavyrulewidth]
    \end{tabular}
    }
    \caption{Performance on the multilingual and cross-lingual STEL-or-content benchmarks, averaged across languages for each style feature. \textsc{mStyleDistance} leads in cross-lingual and overall performance.}
    \label{table:steleval}
    \end{table*}
}

\newcommand{\stelevalfull}{
    \begin{table*}[!h]
    \centering
    \small
    \begin{tabularx}{\textwidth}{l *{5}{>{\centering\arraybackslash}X}}
    \toprule[\heavyrulewidth]
    \textbf{Language} & \textbf{\citet{styleemb}} & \textbf{\textsc{StyleDistance}} & \textbf{xlm-roberta-base} & \textbf{LISA} & \textbf{\textsc{mStyleDistance}} \\
    \midrule
    \multicolumn{6}{l}{\textbf{Simplicity}} \\ \midrule
    de      & 0.23   & 0.06   & 0.00   & 0.00   & \textbf{0.24}    \\
    en      & 0.26   & 0.32   & 0.05   & 0.00   & \textbf{0.36}    \\
    fr      & 0.29   & 0.33   & 0.22   & 0.12   & \textbf{0.46}    \\
    it      & 0.21   & 0.15   & 0.08   & 0.03   & \textbf{0.48}    \\
    ja      & 0.09   & 0.05   & 0.01   & \textbf{0.48}   & 0.14    \\
    pt-br   & 0.10   & 0.07   & 0.04   & 0.03   & \textbf{0.15}    \\
    ru      & 0.26   & 0.24   & 0.07   & 0.15   & \textbf{0.38}    \\
    sl      & 0.43   & 0.43   & 0.46   & 0.39   & \textbf{0.69}    \\
    \midrule
    average & 0.23   & 0.21   & 0.12   & 0.15   & \textbf{0.36}    \\
    \midrule
    \multicolumn{6}{l}{\textbf{Formality}} \\ \midrule
    fr      & 0.70   & 0.81   & 0.16   & 0.06   & \textbf{0.82}    \\
    it      & 0.64   & 0.63   & 0.18   & 0.10   & \textbf{0.69}    \\
    pt-br   & 0.56   & 0.57   & 0.15   & 0.11   & \textbf{0.62}    \\
    \midrule
    average & 0.63   & 0.67   & 0.16   & 0.09   & \textbf{0.71}    \\
    \midrule
    \multicolumn{6}{l}{\textbf{Toxicity}} \\ \midrule
    am      & 0.35   & 0.29   & 0.24   & 0.21   & \textbf{0.53}    \\
    ar      & 0.05   & 0.04   & 0.02   & 0.10   & \textbf{0.18}    \\
    de      & 0.01   & 0.02   & 0.01   & 0.00   & \textbf{0.28}    \\
    en      & \textbf{0.56}   & 0.48   & 0.09   & 0.08   & 0.51    \\
    es      & 0.26   & 0.20   & 0.13   & 0.05   & \textbf{0.35}    \\
    hi      & 0.15   & 0.09   & 0.09   & 0.15   & \textbf{0.37}    \\
    ru      & 0.18   & 0.16   & 0.13   & 0.09   & \textbf{0.61}    \\
    uk      & 0.07   & 0.05   & 0.04   & 0.02   & \textbf{0.25}    \\
    zh      & 0.05   & 0.02   & 0.04   & 0.07   & \textbf{0.23}    \\
    \midrule
    average & 0.19   & 0.15   & 0.09   & 0.09   & \textbf{0.37}    \\
    \midrule
    \multicolumn{6}{l}{\textbf{Positivity}} \\ \midrule
    bn      & 0.27   & 0.13   & 0.04   & 0.23   & \textbf{0.32}    \\
    en      & \textbf{0.21}   & 0.20   & 0.03   & 0.19   & 0.18    \\
    hi      & 0.11   & 0.10   & 0.04   & 0.14   & \textbf{0.22}    \\
    mag     & 0.09   & 0.08   & 0.08   & 0.13   & \textbf{0.41}    \\
    ml      & 0.32   & 0.28   & 0.10   & 0.27   & \textbf{0.39}    \\
    mr      & 0.19   & 0.18   & 0.03   & 0.17   & \textbf{0.22}     \\
    or      & 0.27   & 0.19   & 0.08   & 0.24   & \textbf{0.35}    \\
    pa      & 0.18   & 0.15   & 0.06   & 0.17   & \textbf{0.23}    \\
    te      & 0.39   & 0.34   & 0.20   & 0.29   & \textbf{0.40}    \\
    ur      & 0.24   & 0.20   & 0.08   & 0.28   & \textbf{0.26}    \\
    \midrule
    average & 0.23   & 0.18   & 0.07   & 0.21   & \textbf{0.30}    \\
    \midrule
    \multicolumn{6}{l}{\textbf{Formality (cross-lingual)}} \\ \midrule
    fr-it  & 0.47   & 0.51   & 0.22   & 0.28   & \textbf{0.53}    \\
    fr-pt  & 0.45   & 0.48   & 0.19   & 0.29   & \textbf{0.52}    \\
    it-fr  & 0.48   & \textbf{0.53}   & 0.18   & 0.26   & \textbf{0.53}    \\
    it-pt  & 0.41   & 0.45   & 0.19   & 0.27   & \textbf{0.52}    \\
    pt-fr  & 0.46   & \textbf{0.53}   & 0.17   & 0.27   & \textbf{0.53}    \\
    pt-it  & 0.42   & 0.47   & 0.21   & 0.27   & \textbf{0.52}    \\
    \midrule
    average & 0.45   & 0.49   & 0.19   & 0.27   & \textbf{0.53}    \\
    \bottomrule
    \end{tabularx}
    \caption{Full performance on the multilingual and cross-lingual STEL-or-content benchmarks. For the cross-lingual SoC evaluation, "a-b" means that the anchor sentences were all in language a and alternative sentences were all in language b. \textsc{mStyleDistance} leads in cross-lingual and overall performance.}
    \end{table*}
}

\newcommand{\ablationtable}{
    \begin{table*}[!h]
    \centering
    \small
    \setlength{\tabcolsep}{3pt}
    \scalebox{1.05}{
    \begin{tabular}{l|cccc|c|cc}
    \toprule[\heavyrulewidth]
    & \multicolumn{4}{c}{{\bf Multi-lingual SoC}} & {\bf Cross-lingual SoC} & \\ \midrule
    \textbf{Features Tested} 
    & \textbf{Simplicity} 
    & \textbf{Formality} 
    & \textbf{Toxicity} 
    & \textbf{Positivity} 
    & \textbf{Formality} 
    & \multicolumn{2}{c}{\textbf{Retained Perf (\%)}} \\ 
    & & & & & &  ~~~~ m ~~~~ & c \\ \midrule
    In-Domain
    & 0.36
    & 0.71
    & 0.37
    & 0.30
    & 0.53
    & 100\% & 100\%  \\
    Out of Domain
    & 0.29
    & 0.63
    & 0.31
    & 0.23
    & 0.44
    & 75\% & 74\% \\
    Out of Distribution
    & 0.33
    & 0.39
    & 0.26
    & 0.32
    & 0.40
    & 75\% & 62\% \\
    No Language Overlap
    & 0.27
    & 0.51
    & 0.41
    & 0.32
    & 0.52
    & 89\% & 97\% \\
    \bottomrule[\heavyrulewidth]
    \end{tabular}
    }
    \caption{Full results of the ablation study for \textsc{mStyleDistance} embeddings on the SoC benchmarks.}
    \label{table:ablationeval}
    \end{table*}
}

\newcommand{\ablationsimpletable}{
    \begin{table}[!h]
    \centering
    \small
    \setlength{\tabcolsep}{3pt}
    \begin{tabular}{l|c|c|c|c}
    \toprule[\heavyrulewidth]
     \multirow{2}{*}{\textbf{Features Tested}}
    & \multirow{2}{*}{\textbf{m avg}}
    & \multirow{2}{*}{\textbf{c avg}}
    &  \multicolumn{2}{c}{\textbf{Retained Perf (\%)}} \\ 
    & & & ~~~~ m ~~~~ & c \\ \midrule
    In-Domain
    & 0.38
    & 0.53
    & 100\% & 100\% \\
    Out of Domain
    & 0.31
    & 0.44
    & 75\% & 74\% \\
    Out of Distribution
    & 0.31
    & 0.40
    & 75\% & 62\% \\
    No Language Overlap
    & 0.35
    & 0.52
    & 89\% & 97\% \\
    \bottomrule[\heavyrulewidth]
    \end{tabular}
    \caption{\textsc{mStyleDistance} embeddings under three generalization conditions on the multilingual (m avg) and cross-lingual (c avg) STEL-or-Content tasks.}
    \label{table:simplifiedablation}
    \end{table}
}

\newcommand{\stylefeaturestable}{
    {
    \tiny
    \renewcommand{\arraystretch}{2} % Adds space between rows
    \begin{longtable}{p{3cm} p{3cm} p{6.5cm} p{2cm}}
      \toprule
      \textbf{Style Feature} & \textbf{Positive and Negative Prompts} & \textbf{Style Feature Definition} & \textbf{Excluded In} \\
      \midrule
      \endfirsthead
    
      \toprule
      \textbf{Style Feature Name} & \textbf{Positive and Negative Prompts} & \textbf{Style Feature Definition} & \textbf{Excluded In} \\
      \midrule
      \endhead
    
      \bottomrule
      \endfoot
    
      \bottomrule
      \addlinespace
      \caption{40 style features, with the `Excluded in' column indicating that a particular feature was omitted from our dataset due to its inapplicability to a specific language.}
      \label{table:stylefeaturestable} \\
      \endlastfoot

Usage of Conjunctions & Positive: With conjunctions \newline Negative: Less frequent conjunctions & The "Usage of Conjunctions" text style feature refers to the use of words that connect clauses or sentences. Conjunctions are words like "and", "but", "or", "so", "because", etc. They are used to make sentences longer, more complex, or to show the relationship between different parts of a sentence. & \\
Usage of Numerical Substitution & Positive: With number substitution \newline Negative: Without number substitution & Numerical substitution refers to the practice of replacing certain letters in words with numbers that visually resemble those letters. For example, replacing the letter 'e' with the number '3' in the word 'hello' to make it 'h3llo'. This is a common feature in internet slang and informal digital communication. & Arabic, Hindi, Japanese, Korean, Chinese \\
Usage of Words Indicating Affective Processes & Positive: Affective processes \newline Negative: Cognitive processes & The text style feature "Usage of Words Indicating Affective Processes" refers to the use of words that express emotions, feelings, or attitudes. These could be words that show happiness, sadness, anger, fear, surprise, or any other emotional state. The presence of such words in a text indicates that the writer is expressing some form of emotional reaction or sentiment. & \\
Usage of Metaphors & Positive: With metaphor \newline Negative: Without metaphor & The "Usage of Metaphors" text style feature refers to the presence of phrases or sentences in the text that describe something by comparing it indirectly to something else. This is often done to make a description more vivid or to explain complex ideas in a more understandable way. For example, saying "time is a thief" is a metaphor because it's not literally true but it helps to convey the idea that time passes quickly and can't be regained. & \\
Usage of Long Words & Positive: Long average word length \newline Negative: Short average word length & The "Usage of Long Words" text style feature refers to the frequency or prevalence of long words, typically those with more than six or seven letters, in a given text. This style feature is often used to measure the complexity or sophistication of the text. If a text has many long words, it is said to have a high usage of long words. & Arabic, Japanese, Korean, Chinese\\
Usage of Uppercase Letters & Positive: With uppercase letters \newline Negative: Without uppercase letters & The usage of uppercase letters as a text style feature refers to the frequency or manner in which capital letters are used in a text. This could be for emphasis, to denote shouting or strong emotions, or to highlight specific words or phrases. It's not just about the start of sentences or proper nouns, but also about other uses of capital letters in the text. & Arabic, Hindi, Japanese, Korean, Chinese\\
Usage of Articles & Positive: With articles \newline Negative: Less frequent articles & The "Usage of Articles" text style feature refers to how often a text uses words like "a", "an", and "the". These words are called articles and they are used before nouns. This feature measures the frequency of these articles in a given text. & Arabic, Hindi, Japanese, Korean, Russian, Chinese \\
Usage of Text Emojis & Positive: Text Emojis \newline Negative: No Emojis & The text style feature "Usage of Text Emojis" refers to the inclusion of emoticons or smileys in the text. These are combinations of keyboard characters that represent facial expressions or emotions, such as :-D for a big grin or happy face. The presence of these symbols in a text indicates the use of this style feature. & \\
Usage of Nominalizations & Positive: With nominalizations \newline Negative: Without nominalizations & Nominalizations refer to the use of verbs, adjectives, or adverbs as nouns in a sentence. This style feature is often used to make sentences more concise or formal. For example, "the investigation of the crime" is a nominalization of "investigate the crime". & \\
Frequent Usage of Function Words & Positive: With function words \newline Negative: Less frequent function words & The text style feature "Frequent Usage of Function Words" refers to the regular use of words that have little meaning on their own but work in combination with other words to express grammatical relationships. These words include prepositions (like 'in', 'at', 'on'), conjunctions (like 'and', 'but', 'or'), articles (like 'a', 'an', 'the'), and pronouns (like 'he', 'they', 'it'). & \\
Usage of Self-Focused Perspective or Words & Positive: Self-focused \newline Negative: Third-person singular & The "Usage of Self-Focused Perspective or Words" text style feature refers to the use of words or phrases that focus on the speaker or writer themselves. This includes the use of first-person pronouns like "I", "me", "my", "mine", and "myself", or statements that express the speaker's personal thoughts, feelings, or experiences. & \\
Usage of Formal Tone & Positive: Formal \newline Negative: Informal & The "Usage of Formal Tone" text style feature refers to the use of language that is polite, impersonal and adheres to established conventions in grammar and syntax. It avoids slang, contractions, colloquialisms, and often uses more complex sentence structures. This style is typically used in professional, academic, or official communications.&  \\
Usage of Emojis & Positive: With Emojis \newline Negative: No Emojis & The "Usage of Emojis" text style feature refers to the inclusion of emojis, or digital icons, in a text. Emojis are often used to express emotions, ideas, or objects without using words. If a text contains emojis, it has this style feature. & \\
Usage of Offensive Language & Positive: Offensive \newline Negative: Non-Offensive & The "Usage of Offensive Language" text style feature refers to the presence of words or phrases in the text that are considered rude, disrespectful, or inappropriate. These can include swear words, slurs, or any language that could be seen as insulting or derogatory. & \\
Usage of Present Tense and Present-Focused Words & Positive: Present-focused \newline Negative: Future-focused & The text style feature "Usage of Present Tense and Present-Focused Words" refers to the use of verbs in the present tense and words that focus on the current moment or situation. This means the text is primarily discussing events, actions, or states that are happening now or general truths. It's like the text is talking about what is happening in the present time. & \\
Presence of Misspelled Words & Positive: Sentence With a Few Misspelled Words \newline Negative: Normal Sentence & The text style feature "Presence of Misspelled Words" refers to the occurrence of words in a text that are not spelled correctly according to standard dictionary spelling. This could be due to typing errors, lack of knowledge about the correct spelling, or intentional for stylistic or informal communication purposes. & \\
Incorporation of Humor & Positive: With Humor \newline Negative: Without Humor & The "Incorporation of Humor" text style feature refers to the use of language, phrases, or expressions in a text that are intended to make the reader laugh or feel amused. This could include jokes, puns, funny anecdotes, or witty remarks. It's all about adding a touch of comedy or light-heartedness to the text. & \\
Usage of Personal Pronouns & Positive: With personal pronouns \newline Negative: Less frequent pronouns & The "Usage of Personal Pronouns" text style feature refers to the use of words in a text that refer to a specific person or group of people. These words include "I", "you", "he", "she", "it", "we", and "they". The presence of these words in a text can indicate a more personal or direct style of communication. & \\
Fluency in Sentence Construction & Positive: Fluent sentence \newline Negative: Disfluent sentence & "Fluency in Sentence Construction" refers to the smoothness and ease with which sentences are formed and flow together. It involves using correct grammar, appropriate vocabulary, and logical connections between ideas. A text with this feature would read smoothly, without abrupt changes or awkward phrasing. & \\
Usage of Only Uppercase Letters & Positive: All Upper Case \newline Negative: Proper Capitalization & The usage of only uppercase letters style feature refers to the practice of writing all the letters in a text in capital letters. This means that every single letter in the text, whether at the beginning, middle, or end of a sentence, is capitalized. It's like the 'Caps Lock' key on your keyboard is always turned on while typing the text. & Arabic, Hindi, Japanese, Korean, Chinese\\
Usage of Self-Focused Perspective or Words & Positive: Self-focused \newline Negative: Inclusive-focused & The "Usage of Self-Focused Perspective or Words" text style feature refers to the use of words or phrases that focus on the speaker or writer themselves. This includes the use of first-person pronouns like "I", "me", "my", "mine", and "myself", or statements that express the speaker's personal thoughts, feelings, or experiences. & \\
Usage of Pronouns & Positive: With pronouns \newline Negative: Less frequent pronouns & The "Usage of Pronouns" text style feature refers to the frequency and types of pronouns used in a text. Pronouns are words like 'he', 'she', 'it', 'they', 'we', 'you', 'I', etc., that stand in place of names or nouns in sentences. This feature can indicate the level of personalization, formality, or perspective in a text. & \\
Usage of Words Indicating Cognitive Processes & Positive: Cognitive process \newline Negative: Perceptual process & The text style feature "Usage of Words Indicating Cognitive Processes" refers to the use of words that show thinking or mental processes. These words can express understanding, knowledge, belief or doubt. For example, words like 'think', 'know', 'believe', 'understand' are used to indicate cognitive processes. & \\
Complex Sentence Structure & Positive: Complex \newline Negative: Simple & The "Complex Sentence Structure" text style feature refers to sentences that contain multiple ideas or points, often connected by conjunctions (like 'and', 'but', 'or') or punctuation (like commas, semicolons). These sentences often include dependent clauses, which are parts of the sentence that can't stand alone as a complete thought, alongside independent clauses, which can stand alone. In simpler terms, if a sentence has more than one part and these parts are linked together in a way that they give more detailed information or express multiple thoughts, it has a complex sentence structure. & \\
Positive Sentiment Expression & Positive: Positive \newline Negative: Negative & Positive Sentiment Expression is a text style feature that refers to the use of words, phrases, or expressions that convey a positive or optimistic viewpoint or emotion. This could include expressions of happiness, joy, excitement, love, or any other positive feelings. The text is considered to have this feature if it makes the reader feel good or positive after reading it. & \\
Usage of Numerical Digits & Positive: With digits \newline Negative: Less frequent digits & The "Usage of Numerical Digits" text style feature refers to the presence and use of numbers in a text. This includes any digit from 0-9 used alone or in combination to represent quantities, dates, times, or any other numerical information. & \\
Usage of Words Indicating Affective Process & Positive: Affective process \newline Negative: Perceptual process & The "Usage of Words Indicating Affective Process" text style feature refers to the use of words that express emotions, feelings, or attitudes. These words can show positive or negative sentiments, like happiness, anger, love, or hate. If a text uses a lot of these words, it means the writer is expressing a lot of emotion or personal feelings. & \\
Usage of Active Voice & Positive: Active \newline Negative: Passive & The usage of active voice in a text style feature refers to sentences where the subject performs the action stated by the verb. In other words, the subject is active and directly involved in the action. For example, in the sentence "The cat chased the mouse", 'the cat' is the subject that is actively doing the chasing. & \\
Usage of Only Lowercase Letters & Positive: All Lower Case \newline Negative: Proper Capitalization & The style feature "usage of only lowercase letters" refers to the practice of writing all words in a text with small letters only, without using any capital letters. This means that even the first word of a sentence, proper nouns, or the pronoun 'I' are not capitalized. It's like writing a whole text without ever pressing the shift key on your keyboard. & Arabic, Hindi, Japanese, Korean, Chinese\\
Frequent Usage of Common Verbs & Positive: With common verbs \newline Negative: Less frequent common verbs & The text style feature "Frequent Usage of Common Verbs" refers to the regular use of basic action words in a text. These are often simple, everyday verbs that are widely used in language, such as 'is', 'have', 'do', 'say', 'go', etc. If a text frequently uses these common verbs, it has this style feature. & \\
Usage of Prepositions & Positive: With prepositions \newline Negative: Less frequent prepositions & The "Usage of Prepositions" text style feature refers to the use of words that link nouns, pronouns, or phrases to other words within a sentence. These words often indicate location, direction, time, or manner. Examples of prepositions include words like "in", "at", "on", "over", "under", "after", and "before". & \\
Usage of Self-Focused Language & Positive: Self-focused \newline Negative: Audience-focused & The "Usage of Self-Focused Language" text style feature refers to the use of words or phrases that focus on the speaker or writer themselves. This includes the use of first-person pronouns like "I", "me", "my", "mine", and "myself". It's a way of writing or speaking where the person is often referring to their own thoughts, feelings, or experiences. & \\
Usage of Certain Tone & Positive: Certain \newline Negative: Uncertain & This text style feature refers to the use of a confident tone in writing, where the author avoids using uncertain words or phrases such as 'I think', 'might', or 'seems'. This results in a text that appears more assertive and sure of the information being presented.  & \\
Usage of Present-Focused Tense and Words & Positive: Present-focused \newline Negative: Past-focused & The "Usage of Present-Focused Tense and Words" text style feature refers to the use of verbs in the present tense and words that focus on the current moment or situation. This means the text is primarily discussing events, actions, or states that are happening right now or generally true.  & \\
Usage of Sarcasm & Positive: With sarcasm \newline Negative: Without sarcasm & The "Usage of Sarcasm" text style feature refers to the presence of statements or expressions in the text that mean the opposite of what they literally say, often used to mock or show irritation. This style is often characterized by irony, ridicule, or mockery, and is used to express contempt or to criticize something or someone in a humorous way.  & \\
Usage of Self-Focused Perspective or Words & Positive: Self-focused \newline Negative: You-focused & The "Usage of Self-Focused Perspective or Words" text style feature refers to the use of words or phrases that focus on the speaker or writer themselves. This includes the use of first-person pronouns like "I", "me", "my", "mine", and "myself", or statements that express the speaker's personal thoughts, feelings, or experiences.  & \\
Frequent Usage of Punctuation & Positive: With frequent punctuation \newline Negative: Less Frequent punctuation & The text style feature "Frequent Usage of Punctuation" refers to the regular and abundant use of punctuation marks such as commas, periods, exclamation points, question marks, etc., in a piece of text. This style feature is present when the writer often uses these symbols to structure their sentences, express emotions, or emphasize certain points.  & \\
Usage of Polite Tone & Positive: Polite \newline Negative: Impolite & The "Usage of Polite Tone" text style feature refers to the use of respectful and considerate language in a text. This can include using words like 'please', 'thank you', or phrases that show deference or respect to the reader. It's about making the text sound courteous and respectful, rather than demanding or rude.  & \\
Usage of Contractions & Positive: With contractions \newline Negative: Without contractions & The "Usage of Contractions" text style feature refers to the use of shortened forms of words or phrases in a text. These are typically formed by omitting certain letters or sounds and replacing them with an apostrophe, such as "don't" for "do not" or "I'm" for "I am". If a text frequently uses such shortened forms, it has this style feature.  & Arabic, Hindi, Japanese, Korean, Russian, Chinese\\
Frequent Usage of Determiners & Positive: With determiners \newline Negative: Less frequent determiners & The text style feature "Frequent Usage of Determiners" refers to the regular use of words that introduce a noun and give information about its quantity, proximity, definiteness, etc. These words include 'the', 'a', 'an', 'this', 'that', 'these', 'those', 'my', 'your', 'his', 'her', 'its', 'our', 'their'. If a text often uses such words, it has this style feature.  & \\
    \end{longtable}
    }
}

\newcommand{\trainingdetailstable}{
    \begin{table}[H]
    \small
    \centering
    \begin{tabular}{m{.2\textwidth}>{\raggedleft\arraybackslash}m{.2\textwidth}}
    \toprule[\heavyrulewidth]
    \textbf{Hyperparameter} & \textbf{Value} \\ \midrule
    Model & \texttt{xlm-roberta-base} \\
    Hardware & 4x or 8x NVIDIA RTX A6000 \\
    Distributed Protocol & PyTorch FSDP \\
    Data Type & \small{\texttt{torch.bfloat16}} \\
    Loss Function & \texttt{TripletLoss} \citep{tripletloss} \\
    Triplet Loss Margin & 0.1 \\
    LoRA \citep{lora} & \small{\texttt{all-linear, r=8, lora\_alpha=8, lora\_dropout=0.0}} \\
    Optimizer & \texttt{adamw\_torch} \\
    Learning Rate & 1e-4 \\
    Weight Decay & 0.01 \\
    Learning Rate Scheduler & \small{\texttt{linear}} \\
    Warmup Steps & 0 \\
    Batch Size & 384 \\
    Train-Validation Split & 90/10\% \\
    Early Stopping Threshold & 0.0 \\
    Early Stopping Patience & 1 epoch \\

    \bottomrule[\heavyrulewidth]
    \end{tabular}
    \caption{Hyperparameters selected for contrastive learning training experiments.}
    \label{table:hyperparamtable}
    \end{table}
}

\newcommand{\avevaltable}{
    \begin{table*}[t!]
    \centering
    \fontsize{9.5}{11}\selectfont
    \setlength{\tabcolsep}{3.1pt}
    \scalebox{0.85}{
    \begin{tabular}{>{\raggedright\arraybackslash}p{4cm}cc|ccc|ccc|cccc}
    \toprule[\heavyrulewidth]
    & \multicolumn{2}{c}{PAN 2013} & \multicolumn{3}{c}{PAN 2014} & \multicolumn{3}{c}{PAN 2015} & \multicolumn{4}{c}{Average} \\ \midrule
       \textbf{Model} & \textbf{Greek} & \textbf{Spanish} & \textbf{Greek} & \textbf{Spanish} & \textbf{Dutch} & \textbf{Greek} & \textbf{Spanish} & \textbf{Dutch} & \textbf{Greek} & \textbf{Spanish} & \textbf{Dutch} & \textbf{Overall} \\ \midrule
    \textbf{\citet{styleemb}} & 0.66 & 0.87 & 0.56 & 0.54 & 0.59 & 0.47 & 0.61 & 0.59 & 0.56 & 0.67 & 0.59 & 0.61 \\
    \textbf{\textsc{StyleDistance}} & 0.61 & 0.62 & 0.48 & 0.51 & 0.65 & 0.47 & 0.73 & 0.59 & 0.52 & 0.62 & \textbf{0.62} & 0.59 \\
    \textbf{LISA} & 0.51 & 0.64 & 0.46 & 0.56 & 0.62 & 0.48 & 0.66 & 0.48 & 0.48 & 0.62 & 0.55 & 0.55 \\
    \textbf{\textsc{mStyleDistance}} & 0.41 & 0.78 & 0.53 & 0.56 & 0.63 & 0.58 & 0.53 & 0.38 & \textbf{0.64} & \textbf{0.73} & 0.60 & \textbf{0.66} \\
    \bottomrule[\heavyrulewidth]
    \end{tabular}}
    \caption{Results on the PAN 2013-2015 AV shared task for Greek, Spanish, and Dutch. We report performance separately on each PAN dataset and average performance across datasets for the same language. We use the standard ROC-AUC metric for AV.}
    \label{table:aveval_table}
    \end{table*}
}

\newcommand{\datasetevaltable}{
    \begin{table*}[!h]
    \renewcommand{\arraystretch}{1.2} % Adjusts row spacing to be slightly tighter
    \centering
    \scriptsize
    \setlength{\tabcolsep}{4pt} % Reduces column spacing
    \resizebox{0.9\textwidth}{!}{ % Scales the table to 90% of the text width
    \begin{tabular}{l|cc|cc|cc|cc}
    \toprule[\heavyrulewidth]
    \textbf{Language} &
    \makecell{\textbf{Baseline} \\ \textbf{Feature} \\ \textbf{Presence}} &
    \makecell{\textbf{Feature} \\ \textbf{Presence}} &
    \makecell{\textbf{Baseline} \\ \textbf{Fluency}} &
    \makecell{\textbf{Fluency}} &
    \makecell{\textbf{Baseline} \\ \textbf{Diversity}} &
    \makecell{\textbf{Diversity}} &
    \makecell{\textbf{Baseline} \\ \textbf{Similarity}} &
    \makecell{\textbf{Similarity}} \\
    \midrule
    ar       & 0.5 & 0.7475 & 0.5 & 0.9526 & 0.8278 & 0.8245 & 0.9232 & 0.9156 \\
    de       & 0.5 & 1.0000 & 0.5 & 0.7708 & 0.8345 & 0.8341 & 0.8799 & 0.9171 \\
    es       & 0.5 & 0.8125 & 0.5 & 0.9853 & 0.8449 & 0.8478 & 0.8567 & 0.9298 \\
    fr       & 0.5 & 0.7391 & 0.5 & 0.9855 & 0.8483 & 0.8404 & 0.8573 & 0.9224 \\
    hi       & 0.5 & 0.7595 & 0.5 & 0.9958 & 0.8588 & 0.8253 & 0.9468 & 0.8903 \\
    ja       & 0.5 & 0.6667 & 0.5 & 0.8889 & 0.8528 & 0.8321 & 0.8514 & 0.8761 \\
    ko       & 0.5 & 0.8000 & 0.5 & 0.8972 & 0.8540 & 0.8214 & 0.8652 & 0.9286 \\
    ru       & 0.5 & 0.8000 & 0.5 & 0.8972 & 0.8542 & 0.8097 & 0.8713 & 0.9171 \\
    zh-hans  & 0.5 & 0.7475 & 0.5 & 0.9526 & 0.8571 & 0.8220 & 0.8729 & 0.9322 \\
    \midrule
    \textbf{Average} & \textbf{0.5} & \textbf{0.7859} & \textbf{0.5} & \textbf{0.9251} & \textbf{0.8480} & \textbf{0.8286} & \textbf{0.8805} & \textbf{0.9144} \\
    \bottomrule[\heavyrulewidth]
    \end{tabular}
    }
    \caption{Human and synthetic evaluations on our synthetic dataset.}
    \label{table:dataseteval}
    \end{table*}
}

\newcommand{\promptexample}{%
  \begin{figure}[tbp]
  \footnotesize
  \rule{\linewidth}{0.2pt}
Generate a pair of active and passive Russian sentences with the following attributes:\\

    \quad 1. Topic: Brake parts and components
    
    \quad 2. Length: 15-20 words
    
    \quad 3. Point of view: second-person
    
    \quad 4. Tense: past
    
    \quad 5. Type of Sentence: Exclamation
     \rule{\linewidth}{0.2pt}
    \caption{Example prompt for generating a pair of sentences in Russian.}
    \label{figure:promptexample}
  \end{figure}
}

\newcommand{\ablationdetailstable}{

\begin{table*}[htbp]
    \centering
    \scalebox{0.75}{
    \begin{tabular}{|c|c|}
        \hline
        \textbf{Ablation Condition} & \textbf{Ablated Features and Languages} \\
        \hline
        \textbf{Out-of-Domain} & 
        \parbox{4cm}{{\bf Ablated Style Features:} \vspace{1cm}} \parbox{12cm}{\vspace{2mm}  Usage of Formal Tone, Usage of Contractions, Usage of Numerical Substitution, Complex Sentence Structure, Usage of Positive Tone, Usage of Offensive Tone, Usage of Polite Tone \vspace{2mm} } \\ \hline
        \textbf{Out-of-Distribution} & 
        \parbox{4cm}{{\bf Ablated Style Features:} \vspace{2.4cm}} \parbox{12cm}{\vspace{2mm} Usage of Formal Tone, Usage of Polite Tone, Fluency in Sentence Construction, Usage of Only Uppercase Letters, Usage of Only Lowercase Letters, Incorporation of Humor,  Usage of Sarcasm, Usage of Contractions, Usage of Numerical Substitution, Usage of Text Emojis, Usage of Emojis, Presence of Misspelled Words, Complex Sentence Structure, Usage of Long Words, Usage of Polite Tone,  Usage of Offensive Tone \vspace{2mm} }
        \\
        \hline
        \textbf{No Language Overlap} & 
        \parbox{4cm}{{\bf Ablated Languages:} \vspace{5mm}} \parbox{12cm}{\vspace{2mm} ar (Arabic), de (German), es (Spanish), fr (French), hi (Hindi), ja (Japanese), ru (Russian) \vspace{2mm} } \\ 
        \hline
    \end{tabular}}
    \caption{Style features and languages ablated for \textbf{Out-of-Domain}, \textbf{Out-of-Distribution}, and \textbf{No Language Overlap}, the three ablation conditions in our ablation study.}
    \label{table:ablationdetailsbis}
\end{table*}

}
\usepackage{fancyvrb} 
\usepackage{varwidth}

\usepackage[preprint]{acl}

\usepackage{times}
\usepackage{latexsym}
\usepackage{arydshln}

\usepackage{multirow}
\usepackage[T1]{fontenc}
\usepackage[utf8]{inputenc}

\usepackage{microtype}

\usepackage{inconsolata}

\usepackage{graphicx}
\usepackage{subcaption}

\title{\textsc{mStyleDistance}: Multilingual Style Embeddings and their Evaluation}

\author{%
  Justin Qiu\thanks{Denotes equal contribution}~~~~~~~~~Jiacheng Zhu\footnotemark[1]~~~~~~~~~Ajay Patel~~~~~~~~~   \\\textbf{Marianna Apidianaki~~~~~~~~~Chris Callison-Burch} \\
  University of Pennsylvania\\
  \texttt{\{jsq, jiachzhu, ajayp, marapi, ccb\}@seas.upenn.edu}
}
\begin{document}
\maketitle

% Main sections
\begin{abstract}
Style embeddings are useful for stylistic analysis and style transfer; however, only English style embeddings have been made available. We introduce Multilingual \textsc{StyleDistance} (\textsc{mStyleDistance}), a multilingual style embedding model trained using synthetic data and contrastive learning. We train the model on data from nine languages and create a multilingual STEL-or-Content benchmark \citep{styleemb} that serves to assess the embeddings' quality. We also employ our embeddings in an authorship verification task involving different languages. Our results show that \textsc{mStyleDistance} embeddings outperform existing models on these multilingual style benchmarks and generalize well to unseen features and languages. We make our model publicly available at \url{https://huggingface.co/StyleDistance/mstyledistance}.
\end{abstract}

\section{Introduction}
\label{sec:introduction}

\mainfig

Style embedding models seek to embed texts with similar style closer in the embedding space regardless of their content. Style embeddings are useful for tasks like style transfer and authorship attribution, but only exist for English \citep{styleemb,patel2024styledistancestrongercontentindependentstyle}. Multilingual style embeddings could also serve to automatically evaluate style preservation in machine translation. Models like XLM-RoBERTa \citep{Conneau2019UnsupervisedCR} and E5 \citep{wang2024multilinguale5textembeddings} create multilingual representations for semantic tasks, but have not addressed style mainly due to the scarcity of style datasets.

% We leverage the methodology of \textsc{StyleDistance} of \citet{patel2024styledistancestrongercontentindependentstyle} and extend it to the multilingual setting. % by using the same synthetic procedure used to train English style embeddings.
% Current style embedding methods like Wegmann \citep{styleemb} are typically trained on large web datasets. However, these datasets lack quality content control and use authorship as a proxy for style, which can be inaccurate. Truly parallel style datasets are rare, even for the English language. The problem is magnified in the multilingual setting, particularly for low-resource languages. To address this, /citet{patel2024styledistancestrongercontentindependentstyle} introduced the use of synthetic parallel style examples to train \textsc{StyleDistance}, a style embedding model. However, their work is only done on English. Given that one of the primary benefits of synthetic data is that it isn't reliant on having large amounts of data, we believe this approach can be valuable in the multilingual domain, particularly for low-resource languages. 

%Human-annotated multilingual % parallel style datasets are typically small and not scalable \citep{mukherjee2024multilingualtextstyletransfer, briakou-etal-2021-ola, dementieva2024overview, ryan-etal-2023-revisiting}. 
% Our proposed methodology allows for the creation of multilingual style embeddings based on synthetic data. 

We propose a procedure, called multilingual \textsc{StyleDistance} (\textsc{mStyleDistance}), to train style embeddings using contrastive learning with synthetic data in multiple languages. Early work on style representations learning often involved unlabeled social media data \citep{deepstyle, styleemb, luar, lisa}, but \citet{patel2024styledistancestrongercontentindependentstyle} showed that a contrastive learning objective with synthetic examples (sentence pairs with similar content and different style) can generate high quality style representations for English. 
% \citet{lai-etal-2022-multilingual} used machine-translated parallel data for formality transfer. % focusing on formality
% \citet{krishna-etal-2022-shot} also proposed a few-shot formality transfer method which controls for style with vectors that model stylistic differences between paraphrases. These vectors are different from our embeddings which map all sentences onto a vector space, %Also, both of these works focus on formality, while we 
% and address a wider range of style features. %and other work addressing the lack of parallel style data is sparse.
We create \textsc{mSynthStel}, a synthetic dataset of paraphrases addressing various style features in nine languages, and use it to create our multilingual style embeddings.

In order to evaluate their quality, we contribute a new multilingual and cross-lingual STEL-or-Content (SoC) evaluation benchmark which, following the original SoC evaluation task \cite{styleemb}, measures the ability of a model to embed sentences with the same style closer in the embedding space than sentences with the same content. We show that \textsc{mStyleDistance} embeddings outperform other representations on these evaluations, and demonstrate their usefulness in a downstream setting addressing a multilingual authorship verification task. We publicly release our model, data, and evaluation benchmarks. 

% We create a dataset of parallel positive/negative style examples (near-exact paraphrases) covering up to 40 style features in 9 languages (called \textsc{msynthstel}). We fine-tune a model on this dataset using contrastive learning to create the first publicly available multilingual style embedding model. We evaluate the quality of the obtained style representations on the Multilingual and Crosslingual STEL-or-Content benchmarks, also created by us. These benchmarks extend the STEL-or-Content evaluation for English \citep{styleemb} in the multilingual setting and are aimed to measure the content-independence of multilingual style embeddings. Additionally, we test the usefulness of the multilingual representations in a downstream setting, addressing a multilingual authorship verification task. % Our model is the first publicly available multilingual style embedding model. %similar to available English style embedding models like Wegmann's style embeddings \citep{styleemb} and \textsc{StyleDistance} \citep{patel2024styledistancestrongercontentindependentstyle}.

% Marianna: there are no other multilingual models to compare with

% \input{./sections/2-RelatedWork}
\section{Multilingual Synthetic Data}

\label{sec:datageneration}
%The core component of \textsc{Multilingual StyleDistance} 
%We build on the work of \citet{patel2024styledistancestrongercontentindependentstyle} %by expanding their synthetic dataset.
%who created a synthetic dataset of English sentence pairs that convey similar content (are near paraphrases) but differ in style.
%\paragraph{Languages and Features.}
We extend the \citet{patel2024styledistancestrongercontentindependentstyle} dataset to nine languages ($L$): Arabic, German, Spanish, French, Hindi, Japanese, Korean, Russian, and Chinese.\footnote{These languages were chosen because they align with the linguistic background of our annotators.} %We rely on the style features identified by \citet{patel2024styledistancestrongercontentindependentstyle} excluding some features that are not applicable to some of the target languages. 
%In this section, we present our synthetic data generation process and the evaluation methods used to assess the dataset’s quality. % before employing it to train our style embeddings.
\paragraph{Style Feature Selection}
We use the set of 40 features addressed in \citet{patel2024styledistancestrongercontentindependentstyle}, leaving out features not applicable to specific languages. For example, articles are not relevant for Chinese and Japanese so the corresponding features have been discarded. % are not addressed in these languages. 
Our set of features ($F$) includes syntactic features (e.g., active/passive voice, contractions, frequent use of function words), emotional and cognitive features (e.g., words indicating sentiment or cognitive processes), stylistic and aesthetic features (e.g., metaphors, formal tone), social and interpersonal features (e.g., polite or offensive tone), graphical and digital features (e.g., capitalization, emojis, numerical digits), temporal and aspectual features (e.g., focus on present or future).\footnote{A full list of features and details is given in 
Appendix \ref{sec:appendix:stylefeatures}.} 

\paragraph{Data Generation} For each retained feature $f \in F$ for a language $l \in L$, we generate 100 pairs of positive ({\tt pos}) and negative ({\tt neg}) examples (paraphrases). In each pair, {\tt pos} is a sentence that contains the style feature (e.g., a formal sentence or a metaphorical one) while {\tt neg} does not. This is illustrated in Figure \ref{fig:main} for the feature ``Active Voice''. Features that cannot be removed completely (e.g., ``usage of articles'') are present with higher frequency in {\tt pos} than in {\tt neg} examples.

Using the same prompting workflow as \citet{patel2024styledistancestrongercontentindependentstyle}, we generate sentence pairs by prompting GPT-4 with the DataDreamer library and an attributed prompt \citep{attrprompt} illustrated in Figure \ref{figure:promptexample} \citep{gpt4, datadreamer}. For diversity, a ``Topic'' for each generation is sampled by extracting a random sentence from a random document in the C4 corpus \citep{t5andc4}, and prompting GPT-4 to identify the topic. For further details on all prompts, see Appendix \ref{sec:appendix:generationdetails}.

We experiment with two approaches to multilingual data generation. In our first approach, sentence pairs are directly generated in each $l \in L$ using a language-specific instruction in the prompt, as illustrated in Figure \ref{figure:promptexample}. % to generate directly in Russian.
In our second approach, English sentence pairs are generated using the prompting workflow and then translated into each target language $l \in L$.%with MT 
\footnote{We use \href{https://www.deepl.com}{DeepL} for all languages except for Hindi which is not supported, where we use \href{https://translate.google.com}{Google Translate} instead.} %We used both methods for data generation and conducted human validation to determine the best approach for each language.
We generate data using both methods and conduct human validation on a random 10\% split in order to determine the best approach for each language. 
\promptexample
% across all nine target languages 
% and selected the best approach as follows: % the stronger approach % for each language 
%based on a data validation procedure: % described below. %subsequent evaluations (see ~\nameref{sec:datavalidation}). 
% \section{Data Validation}
%\paragraph{Data Validation}
% \label{sec:datavalidation} %We evaluated the quality of our data through both human and automatic evaluations, following \citet{patel2024styledistancestrongercontentindependentstyle}. For the 
%For human validation, we split our dataset into a random 10\% split. 
%We recruited NLP students as annotators \footnote{MS and undergraduate students who are native speakers of a language and were offered extra credit for participation.}. 
%Our annotators\footnote{MS and undergraduate students native speakers of a language, who were offered extra credit for participation.} were shown the sentences in the test split and were asked to (a) judge whether a style feature was present in a sentence (Yes/No/Maybe), and (b) rate its fluency. 
Each sentence pair for a style feature creates two annotation task instances. For each task instance of a given style feature and the positive or negative sentence, we asked annotators\footnote{MS and undergraduate students native speakers of a language, who were offered extra credit for participation.} to provide: 1) a judgment on if the style feature is present in the sentence, and 2) a rating of the fluency.
We only considered instances annotated by at least three annotators. Inter-annotator agreement was $\alpha$ = $0.4247$ \citep{krippendorffalpha}. More details on the annotation can be found in Appendix \ref{sec:appendix:humanannotation}.

We calculate an aggregate ``feature presence'' accuracy score for each $l \in L$ by calculating whether the average feature presence score over all annotations is higher for the positive sentence than for the negative sentence in a pair. \footnote{We assign 1 for ``Yes'', 0.5 for ``Possibly'', and 0 for ``No''.} % 0, 0.5, and 1 for "No", "Maybe", and "Yes", respectively.}. 
We calculate an aggregate fluency score by taking an average of the fluency scores that each annotator gave each text \footnote{ We assign a score of 1 for "Fluent", 0.67 for ``Mostly Fluent'', 0.33 for ``Mostly Disfluent'', and 0 for ``Disfluent''.}. We selected the best approach (direct generation v.s. English \textrightarrow ~MT) for each $l \in L$ as the one that produced the most fluent sentences, or the highest feature presence score if both produced similarly fluent generations. The direct approach was selected for all languages in $L$ except for Japanese and Hindi. Our final average feature presence and fluency scores over all $l \in L$, with the best generation approach selected for each $l$, are 0.79 and 0.93, respectively, both above random chance (0.5). Detailed results by language are given in Appendix \ref{sec:appendix:dataseteval}. 

\paragraph{Data Validation} Following \citet{patel2024styledistancestrongercontentindependentstyle}, we also perform automatic validations of the generated data. We validate whether our parallel examples are indeed paraphrases by computing their average cosine similarity\footnote{We use \texttt{\small paraphrase-multilingual-mpnet-base-v2}.} \cite{sentencetransformers}. For comparison, we calculate the similarity of gold-standard paraphrases in the multilingual dataset compiled by \citet{scherrer-2020-tapaco} \footnote{We sample 100 rows of paraphrases from \citet{scherrer-2020-tapaco}.} for each language. The average similarity of our parallel examples is 0.91 which is comparable to that calculated on the \citet{scherrer-2020-tapaco} natural data (0.88), indicating that our pairs are reasonable paraphrases. We measure topic diversity across generated sentences for a $l \in L$ using the metric proposed by \citet{diversityscore} which relies on cosine distance\footnote{In this case, we only evaluate the {\tt pos} sentence which contains the style feature for each pair.}. For comparison, we also compute the diversity score for texts from \citet{scherrer-2020-tapaco}. Again, the two scores are comparable (0.83 vs. 0.85), showing that our examples approach the diversity of natural data. Detailed results can be found in Appendix \ref{sec:appendix:dataseteval}.

\section{Training \textsc{mStyleDistance}}
\label{sec:styledistance}

Following the contrastive training approach of \citet{patel2024styledistancestrongercontentindependentstyle}, we construct feature-specific triples for each language $l \in L$ which contain: an anchor text ($a$); a text with the same style as $a$ but different content ({\tt pos}); a distractor text ({\tt neg}) which is a paraphrase of $a$ or {\tt pos}, but different in style from $a$. We use the multilingual \texttt{xlm-roberta-base} as our base model and train with a triplet loss \citep{Conneau2019UnsupervisedCR,tripletloss}. We ensure half of our triplets are \textit{cross-lingual}, i.e. the {\tt pos} and {\tt neg} texts are randomly sampled from a different language than the anchor text. Full training details can be found in Appendix \ref{sec:appendix:trainingdetails}.
\section{Evaluation}
\label{sec:evaluation}

\stelevaltable
\avevaltable

\paragraph{STEL-or-Content (SoC) Benchmark}

In order to evaluate our style embeddings, we construct a multilingual version of the SoC benchmark \citep{styleemb}.\footnote{The English SoC benchmark covered formality, complexity, number usage, contraction usage, and emoji usage.} SoC measures the ability of a model to embed sentences with the same style closer in the embedding space than sentences with the same content. We construct our \textbf{multilingual SoC benchmark} by sampling 100 pairs of parallel {\tt pos}-{\tt neg} examples for each language from four ground-truth datasets covering four style features and 22 languages: simplicity \citep{ryan-etal-2023-revisiting}, formality \citep{briakou-etal-2021-ola}, toxicity \citep{dementieva2024overview}, and positivity \citep{mukherjee2024multilingualtextstyletransfer}.\footnote{Combined, these datasets cover the following languages: Amharic, Arabic, Bengali, German, English, Spanish, French, Hindi, Italian, Japanese, Magahi, Malayalam, Marathi, Odia, Punjabi, Portuguese (Brazil), Russian, Slovenian, Telugu, Ukrainian, Urdu, and Chinese.} Each instance in our multilingual SoC benchmark consists of a triplet ($a$, {\tt pos}, {\tt neg}) constructed as explained in Section \ref{sec:styledistance}. However, following \citet{styleemb}, the distractor text in our SoC benchmark is always a paraphrase of {\tt pos}.
% \textcolor{gray}{For each instance of our multilingual SoC benchmark, we take two pairs of parallel examples to get (1) an anchor sentence, (2) a sentence with the same style but different content than the anchor, and (3) a sentence with the same content but different style than the anchor.}
A model tested on this benchmark is expected to embed $a$ and {\tt pos} closer than $a$ and {\tt neg}. We rate a model by computing the percent of instances it achieves this goal for across all instances. We form test instances for each $f \in F$ in a language corresponding to all possible triplets, resulting in 4,950  instances for each language-style combination.

We also construct a \textbf{cross-lingual SoC benchmark} that addresses embeddings' ability to capture style similarity \textit{across languages}. This can be useful, for example, to evaluate style preservation in translations. We construct the benchmark with the XFormal dataset \citep{briakou-etal-2021-ola}, which includes parallel data in French, Italian and Portuguese. %We use a similar formulation as described above to create each instance. However, rather than taking both pairs from the same language, we take the sentence pair (which is (2) and (3) described above) from a different language as the anchor pair. 
We again create triplets as described above, but instead of using {\tt pos} and {\tt neg} texts from the same language as the anchor ($a$), we sample them from a different language than $a$. We end up with 19,800 instances for each style in each language. Appendix \ref{sec:appendix:stelfig} contains illustrative examples from each benchmark.

\paragraph{SoC Evaluation Results}

The results obtained by \textsc{mStyleDistance} on the multilingual and cross-lingual SoC benchmarks are presented in Table \ref{table:steleval}. As no general multilingual style embeddings are currently available, we compare with a base multilingual encoder model \texttt{xlm-roberta-base} \citep{Conneau2019UnsupervisedCR} as well as a number of English-trained style embedding models applied in zero-shot fashion to multilingual text: \citet{styleemb}, \textsc{StyleDistance} embeddings \citep{patel2024styledistancestrongercontentindependentstyle}, and \texttt{LISA} \citep{lisa}. 
%\citep{lisa} as baseline models to compare against. 
\textsc{mStyleDistance} embeddings outperform these models on multilingual and cross-lingual SoC tasks confirming its suitability for multilingual applications. The other models perform slightly better than the untrained \texttt{xlm-roberta-base} but still worse than \textsc{mStyleDistance}. 

\paragraph{Ablation Experiments}

\ablationsimpletable

Following \citep{patel2024styledistancestrongercontentindependentstyle}, we run several ablation experiments to evaluate how well our model generalizes to unseen style features and languages. In the \textbf{In-Domain} condition, all style features are included in the training data for every language. To test generalization to unseen style features, in the \textbf{Out of Domain} condition, any style feature directly equivalent to those features tested in the Multilingual and Cross-lingual SoC  benchmarks are excluded from the training data. \textbf{Out of Distribution} further removes any style features indirectly similar or related to those tested in the benchmarks. Finally, \textbf{No Language Overlap} removes the languages present in the benchmark from the training data, in order to test generalization to new languages. Our results are given in Table \ref{table:simplifiedablation} where we measure how much of the performance increase on SoC benchmarks over the base model is retained, despite ablating training data. The results indicate that our method generalizes reasonably well to both ``out of domain'' and ``out of distribution'' style features, and very well to languages not in the training data. Further details on features and languages ablated and full results are provided in Appendices  \ref{sec:appendix:ablationdetails} and \ref{sec:appendix:ablationfull}.

\paragraph{Downstream Task}

Following \citet{patel2024styledistancestrongercontentindependentstyle}, we also evaluate our \textsc{mStyleDistance} embeddings in the authorship verification (AV) task, where the goal is to determine if two documents were written by the same author using stylistic features \citep{authorshipverification}. We use the datasets released by PAN\footnote{\url{https://pan.webis.de}} between 2013 and 2015 in Greek, Spanish, and Dutch. Our results are given in Table \ref{table:aveval_table}. \textsc{mStyleDistance} vectors outperform existing English style embedding models on Spanish and Greek, while Dutch shows similar performance to English \textsc{StyleDistance}. We hypothesize that the linguistic proximity (West Germanic roots) of the two languages helps \textsc{StyleDistance} to generalize to Dutch.

\section{Conclusion}
\label{sec:conclusion}

We introduced a novel approach to learning multilingual style embeddings from synthetic examples, and contribute a two benchmarks for evaluating the quality of multilingual style representations. We show that {\sc mStyleDistance} embeddings are able to distinguish style from content better than other English and multilingual embeddings, and generalize well to unseen features and languages. The authorship verification evaluation shows that {\sc mStyleDistance} embeddings also offer strong performance on multilingual downstream tasks. % We will make our models and datasets publicly available as resources upon publication.
\section * {Limitations}

Our synthetic data generation approaches rely on direct generation or machine translation techniques, both of which have limitations for languages other than English. Most of the languages included in our multilingual and cross-lingual STEL-or-Content and authorship evaluations are not really low-resource, so our evaluations may not  reflect performance in languages  with less resources. Furthermore, our approach  only targets the 33-40 style features (depending on the language) we generated data for, and cannot account for the  wide range of possibilities for style. While these drawbacks may limit our approach, our ablation experiments show strong generalization capabilities to unseen languages and style features indicating promising generalized performance.
\section * {Ethical Considerations}

This work demonstrates the potential of using synthetic data for creating style embeddings in languages lacking such resources, increasing access to broader communities. However, it is important to recognize that the synthetic data generated by large language models may reflect and reinforce existing biases inherent in these models \citep{lisa}. While our approach shows significant promise, ongoing efforts should ensure that such synthetic datasets are evaluated for fairness and bias to promote more equitable outcomes.
\section * {Contribution Statement}

Justin Qiu created the SoC benchmarks and performed most of the evaluations. Jiacheng Zhu created \textsc{mSynthStel} by carrying out data generation and human annotation collection. Ajay Patel trained \textsc{mStyleDistance}. Marianna Apidianaki helped greatly with advising and the final version of the manuscript. Chris Callison-Burch also gave us valuable advice as our advisor. All authors contributed to the final manuscript. 
\section*{Acknowledgements}

This research is supported in part by the Office of the Director of National Intelligence (ODNI), Intelligence Advanced Research Projects Activity (IARPA), via the HIATUS Program contract \#2022-22072200005. The views and conclusions contained herein are those of the authors and should not be interpreted as necessarily representing the official policies, either expressed or implied, of ODNI, IARPA, or the U.S. Government. The U.S. Government is authorized to reproduce and distribute reprints for governmental purposes notwithstanding any copyright annotation therein.

\bibliography{custom}

\begin{thebibliography}{22}
\providecommand{\natexlab}[1]{#1}

\bibitem[{Briakou et~al.(2021)Briakou, Lu, Zhang, and Tetreault}]{briakou-etal-2021-ola}
Eleftheria Briakou, Di~Lu, Ke~Zhang, and Joel Tetreault. 2021.
\newblock \href {https://doi.org/10.18653/v1/2021.naacl-main.256} {Ol{\'a}, bonjour, salve! {XFORMAL}: A benchmark for multilingual formality style transfer}.
\newblock In \emph{Proceedings of the 2021 Conference of the North American Chapter of the Association for Computational Linguistics: Human Language Technologies}, pages 3199--3216, Online. Association for Computational Linguistics.

\bibitem[{Conneau et~al.(2019)Conneau, Khandelwal, Goyal, Chaudhary, Wenzek, Guzm{\'a}n, Grave, Ott, Zettlemoyer, and Stoyanov}]{Conneau2019UnsupervisedCR}
Alexis Conneau, Kartikay Khandelwal, Naman Goyal, Vishrav Chaudhary, Guillaume Wenzek, Francisco Guzm{\'a}n, Edouard Grave, Myle Ott, Luke Zettlemoyer, and Veselin Stoyanov. 2019.
\newblock \href {https://api.semanticscholar.org/CorpusID:207880568} {Unsupervised cross-lingual representation learning at scale}.
\newblock In \emph{Annual Meeting of the Association for Computational Linguistics}.

\bibitem[{Dementieva et~al.(2024)Dementieva, Moskovskiy, Babakov, Ayele, Rizwan, Schneider, Wang, Yimam, Ustalov, Stakovskii, Smirnova, Elnagar, Mukherjee, and Panchenko}]{dementieva2024overview}
Daryna Dementieva, Daniil Moskovskiy, Nikolay Babakov, Abinew~Ali Ayele, Naquee Rizwan, Frolian Schneider, Xintog Wang, Seid~Muhie Yimam, Dmitry Ustalov, Elisei Stakovskii, Alisa Smirnova, Ashraf Elnagar, Animesh Mukherjee, and Alexander Panchenko. 2024.
\newblock Overview of the multilingual text detoxification task at pan 2024.
\newblock In \emph{Working Notes of CLEF 2024 - Conference and Labs of the Evaluation Forum}. CEUR-WS.org.

\bibitem[{Hay et~al.(2020)Hay, Doan, Popineau, and Ait~Elhara}]{deepstyle}
Julien Hay, Bich-Li\^{e}n Doan, Fabrice Popineau, and Ouassim Ait~Elhara. 2020.
\newblock Representation learning of writing style.
\newblock In \emph{Proceedings of the 6th Workshop on Noisy User-generated Text (W-NUT 2020)}.

\bibitem[{Hu et~al.(2021)Hu, Shen, Wallis, Allen-Zhu, Li, Wang, Wang, and Chen}]{lora}
Edward~J. Hu, Yelong Shen, Phillip Wallis, Zeyuan Allen-Zhu, Yuanzhi Li, Shean Wang, Lu~Wang, and Weizhu Chen. 2021.
\newblock \href {https://arxiv.org/abs/2106.09685} {Lora: Low-rank adaptation of large language models}.
\newblock \emph{Preprint}, arXiv:2106.09685.

\bibitem[{Koppel and Winter(2014)}]{authorshipverification}
Moshe Koppel and Yaron Winter. 2014.
\newblock Determining if two documents are written by the same author.
\newblock \emph{Journal of the Association for Information Science and Technology}, 65(1):178--187.

\bibitem[{Krippendorff(2011)}]{krippendorffalpha}
Klaus Krippendorff. 2011.
\newblock \href {https://api.semanticscholar.org/CorpusID:59901023} {Computing krippendorff's alpha-reliability}.

\bibitem[{Mukherjee et~al.(2024)Mukherjee, Ojha, Bansal, Alok, McCrae, and Dušek}]{mukherjee2024multilingualtextstyletransfer}
Sourabrata Mukherjee, Atul~Kr. Ojha, Akanksha Bansal, Deepak Alok, John~P. McCrae, and Ondřej Dušek. 2024.
\newblock \href {https://arxiv.org/abs/2405.20805} {Multilingual text style transfer: Datasets i\& models for indian languages}.
\newblock \emph{Preprint}, arXiv:2405.20805.

\bibitem[{OpenAI et~al.(2024)OpenAI, Achiam, Adler, Agarwal, Ahmad, Akkaya, Aleman, Almeida, Altenschmidt, Altman, Anadkat, Avila, Babuschkin, Balaji, Balcom, Baltescu, Bao, Bavarian, Belgum, Bello, Berdine, Bernadett-Shapiro, Berner, Bogdonoff, Boiko, Boyd, Brakman, Brockman, Brooks, Brundage, Button, Cai, Campbell, Cann, Carey, Carlson, Carmichael, Chan, Chang, Chantzis, Chen, Chen, Chen, Chen, Chen, Chess, Cho, Chu, Chung, Cummings, Currier, Dai, Decareaux, Degry, Deutsch, Deville, Dhar, Dohan, Dowling, Dunning, Ecoffet, Eleti, Eloundou, Farhi, Fedus, Felix, Fishman, Forte, Fulford, Gao, Georges, Gibson, Goel, Gogineni, Goh, Gontijo-Lopes, Gordon, Grafstein, Gray, Greene, Gross, Gu, Guo, Hallacy, Han, Harris, He, Heaton, Heidecke, Hesse, Hickey, Hickey, Hoeschele, Houghton, Hsu, Hu, Hu, Huizinga, Jain, Jain, Jang, Jiang, Jiang, Jin, Jin, Jomoto, Jonn, Jun, Kaftan, Łukasz Kaiser, Kamali, Kanitscheider, Keskar, Khan, Kilpatrick, Kim, Kim, Kim, Kirchner, Kiros, Knight, Kokotajlo, Łukasz Kondraciuk,
  Kondrich, Konstantinidis, Kosic, Krueger, Kuo, Lampe, Lan, Lee, Leike, Leung, Levy, Li, Lim, Lin, Lin, Litwin, Lopez, Lowe, Lue, Makanju, Malfacini, Manning, Markov, Markovski, Martin, Mayer, Mayne, McGrew, McKinney, McLeavey, McMillan, McNeil, Medina, Mehta, Menick, Metz, Mishchenko, Mishkin, Monaco, Morikawa, Mossing, Mu, Murati, Murk, Mély, Nair, Nakano, Nayak, Neelakantan, Ngo, Noh, Ouyang, O'Keefe, Pachocki, Paino, Palermo, Pantuliano, Parascandolo, Parish, Parparita, Passos, Pavlov, Peng, Perelman, de~Avila Belbute~Peres, Petrov, de~Oliveira~Pinto, Michael, Pokorny, Pokrass, Pong, Powell, Power, Power, Proehl, Puri, Radford, Rae, Ramesh, Raymond, Real, Rimbach, Ross, Rotsted, Roussez, Ryder, Saltarelli, Sanders, Santurkar, Sastry, Schmidt, Schnurr, Schulman, Selsam, Sheppard, Sherbakov, Shieh, Shoker, Shyam, Sidor, Sigler, Simens, Sitkin, Slama, Sohl, Sokolowsky, Song, Staudacher, Such, Summers, Sutskever, Tang, Tezak, Thompson, Tillet, Tootoonchian, Tseng, Tuggle, Turley, Tworek, Uribe, Vallone,
  Vijayvergiya, Voss, Wainwright, Wang, Wang, Wang, Ward, Wei, Weinmann, Welihinda, Welinder, Weng, Weng, Wiethoff, Willner, Winter, Wolrich, Wong, Workman, Wu, Wu, Wu, Xiao, Xu, Yoo, Yu, Yuan, Zaremba, Zellers, Zhang, Zhang, Zhao, Zheng, Zhuang, Zhuk, and Zoph}]{gpt4}
OpenAI, Josh Achiam, Steven Adler, Sandhini Agarwal, Lama Ahmad, Ilge Akkaya, Florencia~Leoni Aleman, Diogo Almeida, Janko Altenschmidt, Sam Altman, Shyamal Anadkat, Red Avila, Igor Babuschkin, Suchir Balaji, Valerie Balcom, Paul Baltescu, Haiming Bao, Mohammad Bavarian, Jeff Belgum, Irwan Bello, Jake Berdine, Gabriel Bernadett-Shapiro, Christopher Berner, Lenny Bogdonoff, Oleg Boiko, Madelaine Boyd, Anna-Luisa Brakman, Greg Brockman, Tim Brooks, Miles Brundage, Kevin Button, Trevor Cai, Rosie Campbell, Andrew Cann, Brittany Carey, Chelsea Carlson, Rory Carmichael, Brooke Chan, Che Chang, Fotis Chantzis, Derek Chen, Sully Chen, Ruby Chen, Jason Chen, Mark Chen, Ben Chess, Chester Cho, Casey Chu, Hyung~Won Chung, Dave Cummings, Jeremiah Currier, Yunxing Dai, Cory Decareaux, Thomas Degry, Noah Deutsch, Damien Deville, Arka Dhar, David Dohan, Steve Dowling, Sheila Dunning, Adrien Ecoffet, Atty Eleti, Tyna Eloundou, David Farhi, Liam Fedus, Niko Felix, Simón~Posada Fishman, Juston Forte, Isabella Fulford, Leo
  Gao, Elie Georges, Christian Gibson, Vik Goel, Tarun Gogineni, Gabriel Goh, Rapha Gontijo-Lopes, Jonathan Gordon, Morgan Grafstein, Scott Gray, Ryan Greene, Joshua Gross, Shixiang~Shane Gu, Yufei Guo, Chris Hallacy, Jesse Han, Jeff Harris, Yuchen He, Mike Heaton, Johannes Heidecke, Chris Hesse, Alan Hickey, Wade Hickey, Peter Hoeschele, Brandon Houghton, Kenny Hsu, Shengli Hu, Xin Hu, Joost Huizinga, Shantanu Jain, Shawn Jain, Joanne Jang, Angela Jiang, Roger Jiang, Haozhun Jin, Denny Jin, Shino Jomoto, Billie Jonn, Heewoo Jun, Tomer Kaftan, Łukasz Kaiser, Ali Kamali, Ingmar Kanitscheider, Nitish~Shirish Keskar, Tabarak Khan, Logan Kilpatrick, Jong~Wook Kim, Christina Kim, Yongjik Kim, Jan~Hendrik Kirchner, Jamie Kiros, Matt Knight, Daniel Kokotajlo, Łukasz Kondraciuk, Andrew Kondrich, Aris Konstantinidis, Kyle Kosic, Gretchen Krueger, Vishal Kuo, Michael Lampe, Ikai Lan, Teddy Lee, Jan Leike, Jade Leung, Daniel Levy, Chak~Ming Li, Rachel Lim, Molly Lin, Stephanie Lin, Mateusz Litwin, Theresa Lopez, Ryan
  Lowe, Patricia Lue, Anna Makanju, Kim Malfacini, Sam Manning, Todor Markov, Yaniv Markovski, Bianca Martin, Katie Mayer, Andrew Mayne, Bob McGrew, Scott~Mayer McKinney, Christine McLeavey, Paul McMillan, Jake McNeil, David Medina, Aalok Mehta, Jacob Menick, Luke Metz, Andrey Mishchenko, Pamela Mishkin, Vinnie Monaco, Evan Morikawa, Daniel Mossing, Tong Mu, Mira Murati, Oleg Murk, David Mély, Ashvin Nair, Reiichiro Nakano, Rajeev Nayak, Arvind Neelakantan, Richard Ngo, Hyeonwoo Noh, Long Ouyang, Cullen O'Keefe, Jakub Pachocki, Alex Paino, Joe Palermo, Ashley Pantuliano, Giambattista Parascandolo, Joel Parish, Emy Parparita, Alex Passos, Mikhail Pavlov, Andrew Peng, Adam Perelman, Filipe de~Avila Belbute~Peres, Michael Petrov, Henrique~Ponde de~Oliveira~Pinto, Michael, Pokorny, Michelle Pokrass, Vitchyr~H. Pong, Tolly Powell, Alethea Power, Boris Power, Elizabeth Proehl, Raul Puri, Alec Radford, Jack Rae, Aditya Ramesh, Cameron Raymond, Francis Real, Kendra Rimbach, Carl Ross, Bob Rotsted, Henri Roussez,
  Nick Ryder, Mario Saltarelli, Ted Sanders, Shibani Santurkar, Girish Sastry, Heather Schmidt, David Schnurr, John Schulman, Daniel Selsam, Kyla Sheppard, Toki Sherbakov, Jessica Shieh, Sarah Shoker, Pranav Shyam, Szymon Sidor, Eric Sigler, Maddie Simens, Jordan Sitkin, Katarina Slama, Ian Sohl, Benjamin Sokolowsky, Yang Song, Natalie Staudacher, Felipe~Petroski Such, Natalie Summers, Ilya Sutskever, Jie Tang, Nikolas Tezak, Madeleine~B. Thompson, Phil Tillet, Amin Tootoonchian, Elizabeth Tseng, Preston Tuggle, Nick Turley, Jerry Tworek, Juan Felipe~Cerón Uribe, Andrea Vallone, Arun Vijayvergiya, Chelsea Voss, Carroll Wainwright, Justin~Jay Wang, Alvin Wang, Ben Wang, Jonathan Ward, Jason Wei, CJ~Weinmann, Akila Welihinda, Peter Welinder, Jiayi Weng, Lilian Weng, Matt Wiethoff, Dave Willner, Clemens Winter, Samuel Wolrich, Hannah Wong, Lauren Workman, Sherwin Wu, Jeff Wu, Michael Wu, Kai Xiao, Tao Xu, Sarah Yoo, Kevin Yu, Qiming Yuan, Wojciech Zaremba, Rowan Zellers, Chong Zhang, Marvin Zhang, Shengjia
  Zhao, Tianhao Zheng, Juntang Zhuang, William Zhuk, and Barret Zoph. 2024.
\newblock \href {https://arxiv.org/abs/2303.08774} {Gpt-4 technical report}.
\newblock \emph{Preprint}, arXiv:2303.08774.

\bibitem[{Patel et~al.(2024{\natexlab{a}})Patel, Raffel, and Callison-Burch}]{datadreamer}
Ajay Patel, Colin Raffel, and Chris Callison-Burch. 2024{\natexlab{a}}.
\newblock \href {https://doi.org/10.18653/v1/2024.acl-long.208} {{D}ata{D}reamer: A tool for synthetic data generation and reproducible {LLM} workflows}.
\newblock In \emph{Proceedings of the 62nd Annual Meeting of the Association for Computational Linguistics (Volume 1: Long Papers)}, pages 3781--3799, Bangkok, Thailand. Association for Computational Linguistics.

\bibitem[{Patel et~al.(2023)Patel, Rao, Kothary, McKeown, and Callison-Burch}]{lisa}
Ajay Patel, Delip Rao, Ansh Kothary, Kathleen McKeown, and Chris Callison-Burch. 2023.
\newblock \href {https://doi.org/10.18653/v1/2023.findings-emnlp.1020} {Learning interpretable style embeddings via prompting {LLM}s}.
\newblock In \emph{Findings of the Association for Computational Linguistics: EMNLP 2023}, pages 15270--15290, Singapore. Association for Computational Linguistics.

\bibitem[{Patel et~al.(2024{\natexlab{b}})Patel, Zhu, Qiu, Horvitz, Apidianaki, McKeown, and Callison-Burch}]{patel2024styledistancestrongercontentindependentstyle}
Ajay Patel, Jiacheng Zhu, Justin Qiu, Zachary Horvitz, Marianna Apidianaki, Kathleen McKeown, and Chris Callison-Burch. 2024{\natexlab{b}}.
\newblock \href {https://arxiv.org/abs/2410.12757} {Styledistance: Stronger content-independent style embeddings with synthetic parallel examples}.
\newblock \emph{Preprint}, arXiv:2410.12757.

\bibitem[{Raffel et~al.(2020)Raffel, Shazeer, Roberts, Lee, Narang, Matena, Zhou, Li, and Liu}]{t5andc4}
Colin Raffel, Noam Shazeer, Adam Roberts, Katherine Lee, Sharan Narang, Michael Matena, Yanqi Zhou, Wei Li, and Peter~J Liu. 2020.
\newblock Exploring the limits of transfer learning with a unified text-to-text transformer.
\newblock \emph{Journal of Machine Learning Research}, 21:\mbox{1--6}.

\bibitem[{Reimers and Gurevych(2019)}]{sentencetransformers}
Nils Reimers and Iryna Gurevych. 2019.
\newblock Sentence-bert: Sentence embeddings using siamese bert-networks.
\newblock In \emph{Proceedings of the 2019 Conference on Empirical Methods in Natural Language Processing and the 9th International Joint Conference on Natural Language Processing (EMNLP-IJCNLP)}, pages 3982--3992.

\bibitem[{Rivera-Soto et~al.(2021)Rivera-Soto, Miano, Ordonez, Chen, Khan, Bishop, and Andrews}]{luar}
Rafael~A Rivera-Soto, Olivia~Elizabeth Miano, Juanita Ordonez, Barry~Y Chen, Aleem Khan, Marcus Bishop, and Nicholas Andrews. 2021.
\newblock Learning universal authorship representations.
\newblock In \emph{Proceedings of the 2021 Conference on Empirical Methods in Natural Language Processing}, pages 913--919.

\bibitem[{Ryan et~al.(2023)Ryan, Naous, and Xu}]{ryan-etal-2023-revisiting}
Michael Ryan, Tarek Naous, and Wei Xu. 2023.
\newblock \href {https://aclanthology.org/2023.acl-long.269} {Revisiting non-{E}nglish text simplification: A unified multilingual benchmark}.
\newblock In \emph{Proceedings of the 61st Annual Meeting of the Association for Computational Linguistics (Volume 1: Long Papers)}, pages 4898--4927, Toronto, Canada. Association for Computational Linguistics.

\bibitem[{Scherrer(2020)}]{scherrer-2020-tapaco}
Yves Scherrer. 2020.
\newblock \href {https://aclanthology.org/2020.lrec-1.848/} {{T}a{P}a{C}o: A corpus of sentential paraphrases for 73 languages}.
\newblock In \emph{Proceedings of the Twelfth Language Resources and Evaluation Conference}, pages 6868--6873, Marseille, France. European Language Resources Association.

\bibitem[{Schroff et~al.(2015)Schroff, Kalenichenko, and Philbin}]{tripletloss}
Florian Schroff, Dmitry Kalenichenko, and James Philbin. 2015.
\newblock Facenet: A unified embedding for face recognition and clustering.
\newblock In \emph{Proceedings of the IEEE conference on computer vision and pattern recognition}, pages 815--823.

\bibitem[{Wang et~al.(2024)Wang, Yang, Huang, Yang, Majumder, and Wei}]{wang2024multilinguale5textembeddings}
Liang Wang, Nan Yang, Xiaolong Huang, Linjun Yang, Rangan Majumder, and Furu Wei. 2024.
\newblock \href {https://arxiv.org/abs/2402.05672} {Multilingual e5 text embeddings: A technical report}.
\newblock \emph{Preprint}, arXiv:2402.05672.

\bibitem[{Wegmann et~al.(2022)Wegmann, Schraagen, Nguyen et~al.}]{styleemb}
Anna Wegmann, Marijn Schraagen, Dong Nguyen, et~al. 2022.
\newblock Same author or just same topic? towards content-independent style representations.
\newblock In \emph{Proceedings of the 7th Workshop on Representation Learning for NLP}, page 249. Association for Computational Linguistics.

\bibitem[{Yang et~al.(2024)Yang, Gandhi, Wang, Wu, Yao, Callison-Burch, Gee, and Yatskar}]{diversityscore}
Yue Yang, Mona Gandhi, Yufei Wang, Yifan Wu, Michael~S Yao, Chris Callison-Burch, James~C Gee, and Mark Yatskar. 2024.
\newblock A textbook remedy for domain shifts: Knowledge priors for medical image analysis.
\newblock \emph{arXiv preprint arXiv:2405.14839}.

\bibitem[{Yu et~al.(2023)Yu, Zhuang, Zhang, Meng, Ratner, Krishna, Shen, and Zhang}]{attrprompt}
Yue Yu, Yuchen Zhuang, Jieyu Zhang, Yu~Meng, Alexander~J Ratner, Ranjay Krishna, Jiaming Shen, and Chao Zhang. 2023.
\newblock \href {https://proceedings.neurips.cc/paper_files/paper/2023/file/ae9500c4f5607caf2eff033c67daa9d7-Paper-Datasets_and_Benchmarks.pdf} {Large language model as attributed training data generator: A tale of diversity and bias}.
\newblock In \emph{Advances in Neural Information Processing Systems}, volume~36, pages 55734--55784. Curran Associates, Inc.

\end{thebibliography}

% Appendices
\appendix
\clearpage

\section{Style Features and Definitions}
\label{sec:appendix:stylefeatures}
% We list all style features selected for our synthetic dataset below along with the positive and negative prompts (used for constructing a full prompt for generating positive and negative examples as shown in Appendix \ref{sec:genposandneg}) and definitions (used to help define the style feature to human annotators in the annotation interface in Appendix \ref{sec:appendix:humanannotation}).

The style features addressed in our experiments included most of the %were based on 
40 style features in the %presented in 
\citet{patel2024styledistancestrongercontentindependentstyle} dataset.
In Table \ref{table:stylefeaturestable}, we list the 40 style features with an `Excluded in' column indicating the languages where each feature is not applicable and was therefore omitted from our dataset.

% \newpage
\section{Generation Prompts and Details}
\label{sec:appendix:generationdetails}

Below we detail the structure of our prompts and inference parameters used for two multilingual synthetic data generation methods.

\subsection{Extracting Topics from C4}
\label{sec:extracttopic}

We use the same topic extraction method as \citet{patel2024styledistancestrongercontentindependentstyle}, which is derived from the C4 dataset \citep{t5andc4}, to identify 50,000 topics through zero-shot prompting with GPT-4 \citep{gpt4}. These 50,000 fine-grained, unique topics ensure that each sentence pair has a distinct context across various features and languages. We perform topic sampling with a \texttt{temperature} setting of 1.0 and \texttt{top\_p} = 0.0.

\vspace{1em}
{\small
\begin{verbatim}
What is the fine-grained topic of the following 
text: 
 {sentence} Only return the topic. 
\end{verbatim}}
\vspace{1em}

\noindent The fine-grained topic is used as part of the attributed prompt in Section \ref{sec:genposandneg} to ensure diverse generations.

\subsection{Generating Positive and Negative Example Sentences for Each Style} 

We use the same prompt as \citet{patel2024styledistancestrongercontentindependentstyle} to generate positive and negative example sentences in English. We then translate these sentence pairs into the target languages using the DeepL API. The only exception is Hindi, which we translate using Google Translate API due to DeepL's limited language support.

\subsection{Direct Generating Positive and Negative Example Sentences for Each Style Feature into Target Language}
\label{sec:genposandneg}

For direct generation in target language method (method 2), we use the following prompt with \texttt{temperature} setting of 1.0 and \texttt{top\_p} = 1.0

\vspace{1em}
{\small
\begin{verbatim}
Generate a pair of {target language} sentences with
and without sarcasm with the following attributes:
 1. Topic: {topic}
 2. Length: {sentence_length}
 3. Point of view: {point_of_view}
 4. Tense: {tense}
 5. Type of Sentence: {sentence_type}
 
Ensure that the generated sentences meet the
following conditions:
 1. There is no extra information in one sentence
 that is not in the other. 
 2. The difference between the two sentences is 
 subtle. 
 3. The two sentences have the same length.
 {special_conditions_for_style_feature}
 
Use Format:
 With sarcasm: [sentence in {target language}]
 Without sarcasm: [sentence in {target language}]
 
Your response should only consist of the two sentences,
without quotation marks.
\end{verbatim}}
\vspace{1em}
% \newpage
\section{Human Annotation Details}

\mturkinterfacefig

\datasetevaltable % moved to force up a page

%\subsection{Human Annotation Interface}
\label{sec:appendix:humanannotation}

% We provide an example of a task instance in our annotation interface. Human annotators were asked to rate whether the style feature was present or not in the sentence, with the option to also select "Possibly" if the annotator was unsure (instructed to use sparingly). We provide annotators with a definition of each style feature as well.

We show above in Figure \ref{fig:mturkinterface} an instance from the human annotation interface. We first asked annotators whether a given style feature was present in a sentence in their chosen language. We also provided a definition for each style feature to help annotators in their decision. The annotators then had to rate the fluency of the sentence be selecting one of ``Fluent'', ``Mostly Fluent'', ``Mostly Disfluent'', or ``Disfluent''. %We also provided definitions for each style feature so help annotators in their decision. %could clearly understand them.

% \noindent We used a population of graduate students taking a class on natural language processing as the annotators. Each task instance was annotated by 10 distinct human annotators. We assign a score of 0 to ``No'', 0.5 to ``Possibly'', and 1 to ``Yes''. We average the scores from all 10 annotators assigned to each task instance. We consider to have agreement for a positive example if the average score is >=0.5, and for a negative example if the average score is < 0.5.

% We measure inter-annotator agreement using Krippendorf's Alpha \citep{krippendorffalpha} which indicates moderate agreement of 0.55. As a more easily interpretable measure of agreement between annotators, for each task instance, we also find, on average, around 8 out of the 10 annotators annotated in agreement on whether a style feature was present or not in the text.

Our annotators are undergraduate and graduate students from a NLP class and were offered extra credit for their participation in the study. Each instance was annotated by at least three annotators: three for languages with fewer native speakers such as Arabic and Russian; over ten for languages with a large number of native speakers, such as Chinese. We used Krippendorff’s Alpha \citep{krippendorffalpha} to measure inter-annotator agreement, which indicated moderate agreement of $0.4247 \pm 0.1719$.

% \subsection{Chosen Method for each Target Language}
% \label{sec:appendix:chosenmethod}

% \llmormttable

\section{Dataset Evaluation}
\label{sec:appendix:dataseteval}

In Table \ref{table:dataseteval}, we show the full results of human and synthetic evaluations on our synthetic dataset. Our synthetic dataset is comparable to a reference dataset compiled by \citet{scherrer-2020-tapaco} on feature presence, fluency, diversity, and similarity. Note that baselines shown for feature presence and fluency are just 0.5 to represent random guessing.

% \datasetevaltable

% \newpage
\section{Training Details}
\label{sec:appendix:trainingdetails}

Table \ref{table:hyperparamtable} contains details for our hyperparameters for training. More exact training details can be found in the source code provided in the supplementary materials for this work.

\trainingdetailstable

\ablationdetailstable % moved up to force up a page

\ablationtable % moved up to force up a page
% \newpage
\section{Instances from the Multilingual and Cross-lingual SoC Benchmarks}
\label{sec:appendix:stelfig}

\begin{figure}[htbp]
    \centering
    \begin{subfigure}{0.45\textwidth}
        \centering
        \includegraphics[width=\linewidth]{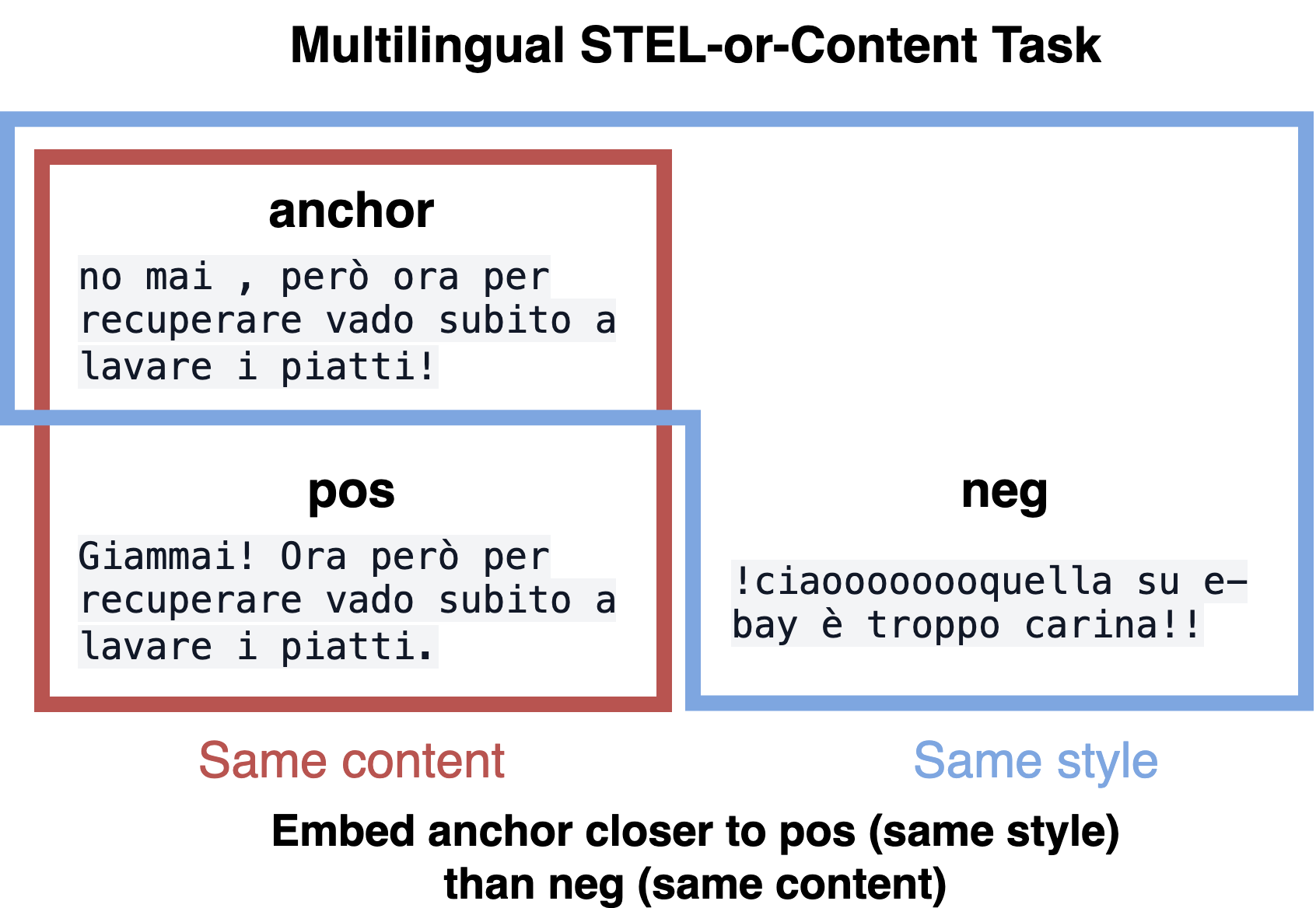}
        % \caption{First figure caption}
        \label{fig:figure1}
    \end{subfigure}
    \hfill
    \begin{subfigure}{0.45\textwidth}
        \centering
        \includegraphics[width=\linewidth]{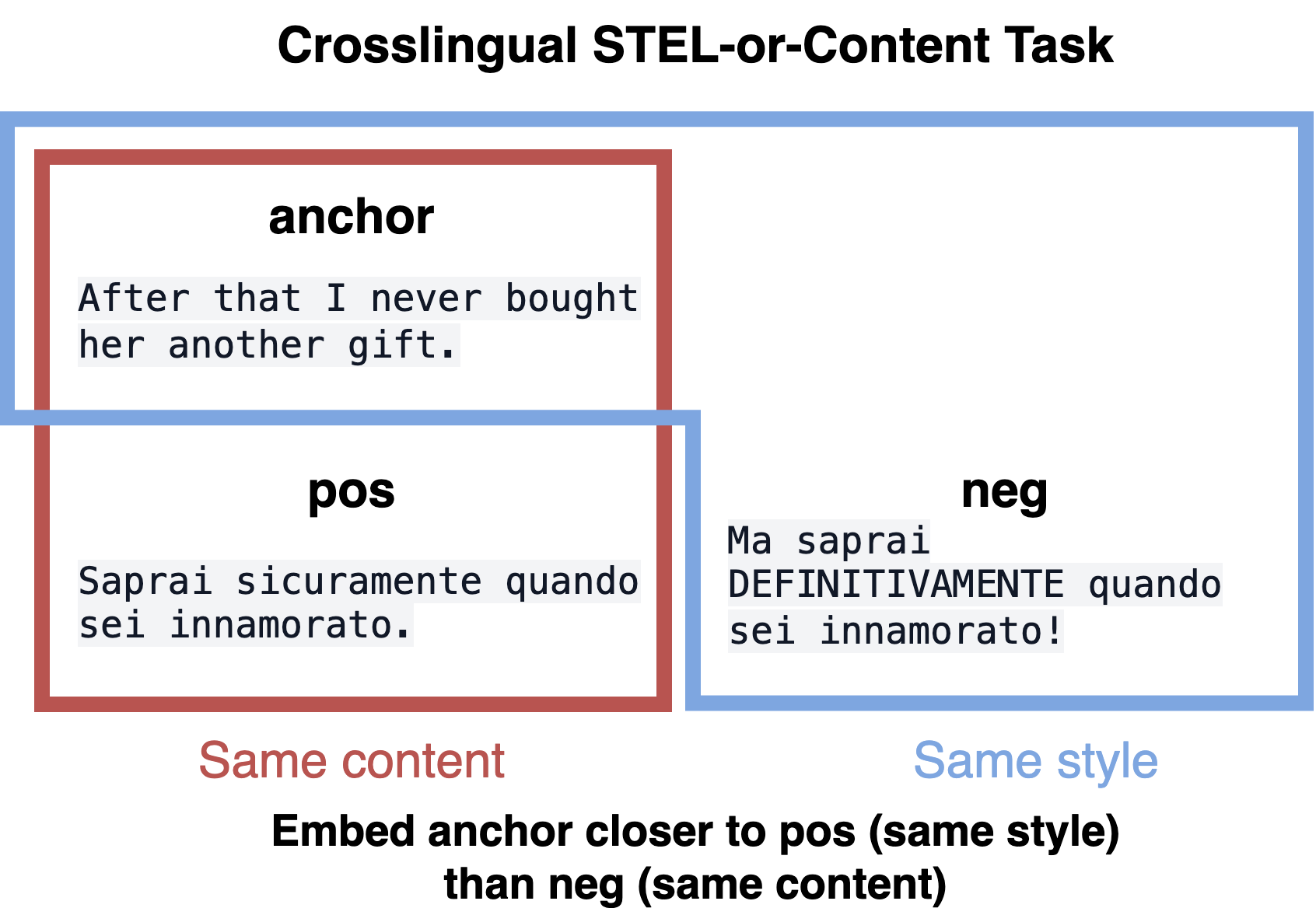}
        % \caption{Second figure caption}
        \label{fig:figure2}
    \end{subfigure}
    \caption{Instances from our multilingual and cross-lingual SoC benchmarks. For multilingual SoC, the anchor is in the same language as the pos and neg sentences. For cross-lingual SoC, the anchor is in a different language from the pos and neg sentences.}
    \label{fig:both_figures}
\end{figure}

In our multilingual SoC benchmark, anchor ($a$) has the same style and different content from a positive example ({\tt pos}), and the same content but different style from a negative example ({\tt neg}). The anchor and the {\tt pos} and {\tt neg} sentences are in the same language. The tested model needs to successfully pair $a$ with {\tt pos} (rather than $a$ and {\tt neg}). Cross-lingual SoC has the same setup as multilingual SoC, except that the {\tt pos} and {\tt neg} examples are in a different language than the anchor. Figure \ref{fig:both_figures} contains instances of each benchmark.

% \multilingualstelorcontentfig

% \noindent Cross-lingual SoC has the same setup as multilingual SoC, except that the {\tt pos} and {\tt neg} examples are in a different language than the anchor. 

% \crosslingualstelorcontentfig

% \newpage
\section{Style Feature and Language Ablation Details}
\label{sec:appendix:ablationdetails}

Full ablation results can be found in Table \ref{table:ablationdetailsbis}.

% \ablationdetailstable
\section{Ablation Full Results}

Table \ref{table:ablationeval} contains full results of the ablation study for \textsc{mStyleDistance} embeddings on the SoC benchmarks under three generalization conditions: Out of Domain, Out of Distribution, and No Language Overlap. For multilingual SoC, we use all four style features: simplicity, formality, toxicity, and positivity. For the cross-lingual variant, we only use formality.

\label{sec:appendix:ablationfull}

% \ablationtable
\onecolumn
\section{Full SoC Results}
\label{sec:appendix:stelfull}

\stelevalfull
\clearpage
\stylefeaturestable

\end{document}